%% file: iccv2023-04349.tex
\documentclass[10pt,twocolumn,letterpaper]{article}

\usepackage{iccv}
\usepackage{times}
\usepackage{epsfig}
\usepackage{graphicx}
\usepackage{amsmath}
\usepackage{amssymb}
\usepackage[accsupp]{axessibility}

\usepackage{multirow}
\usepackage{xcolor}
\usepackage[breaklinks=true,bookmarks=false,colorlinks]{hyperref}

\iccvfinalcopy % *** Uncomment this line for the final submission

\def\iccvPaperID{****} % *** Enter the ICCV Paper ID here

\ificcvfinal\fi

\begin{document}

%%%%%%%%% TITLE
\title{Leveraging Intrinsic Properties for Non-Rigid Garment Alignment}

\author{Siyou Lin\quad Boyao Zhou\quad Zerong Zheng\quad Hongwen Zhang\quad Yebin Liu\\
Tsinghua University\\
Beijing, China\\
{\tt\small {linsy21}@mai1s.tsinghua.edu.cn\quad bzhou22@mail.tsinghua.edu.cn\quad zrzheng1995@foxmail.com}\\
{\tt\small zhanghongwen@tsinghua.edu.cn\quad liuyebin@mail.tsinghua.edu.cn}
}

\maketitle

%%%%%%%%% ABSTRACT
\begin{abstract}
      We address the problem of aligning real-world 3D data of garments, which benefits many applications such as texture learning, physical parameter estimation, generative modeling of garments, etc. Existing extrinsic methods typically perform non-rigid iterative closest point and struggle to align details due to incorrect closest matches and rigidity constraints. While intrinsic methods based on functional maps can produce high-quality correspondences, they work under isometric assumptions and become unreliable for garment deformations which are highly non-isometric. To achieve wrinkle-level as well as texture-level alignment, we present a novel coarse-to-fine two-stage method that leverages intrinsic manifold properties with two neural deformation fields, in the 3D space and the intrinsic space, respectively. The coarse stage performs a 3D fitting, where we leverage intrinsic manifold properties to define a manifold deformation field. The coarse fitting then induces a functional map that produces an alignment of intrinsic embeddings. We further refine the intrinsic alignment with a second neural deformation field for higher accuracy. We evaluate our method with our captured garment dataset, GarmCap. The method achieves accurate wrinkle-level and texture-level alignment and works for difficult garment types such as long coats. Our project page is \href{https://jsnln.github.io/iccv2023_intrinsic/index.html}{https://jsnln.github.io/iccv2023\_intrinsic/index.html}.
\end{abstract}

%%%%%%%%% BODY TEXT
\input{sections/sec1_intro_v7}
\input{sections/sec2_relate_v3}
\input{sections/sec3_method}

\input{sections/sec4_exp}

\input{sections/sec5_conclusion}

\section*{Acknowledgments}
This paper is supported by National Key R\&D Program of China (2022YFF0902200), the NSFC project No.62125107 and No.61827805.

{\small
\bibliographystyle{ieee_fullname}
\bibliography{egbib}
}

\end{document}

%% file: sections/sec1_intro_v7.tex
\section{Introduction}\label{sec:intro}

\input{figs/fig1_tex}

The research into building realistic animatable human avatars has drawn increasing attention in the past few years. Recent developments have demonstrated that compared with full-body avatars~\cite{chen2021snarf,li2023posevocab,lin2022fite,ma2021scale,ma2021pop,saito2021scanimate,zheng2022structured,zheng2023avatarrex}, methods utilizing accurately aligned garment scans to explicitly model garments produce much more realistic geometric deformations and/or rendering results~\cite{feng2022scarf,halimi2022pattern,xiang2022dressing,xiang2021separate}, and enable or improve a number of downstream tasks such as retargeting~\cite{ponsmoll2017clothcap}, texture learning~\cite{xiang2022dressing,xiang2021separate}, pose-driving garment animation~\cite{halimi2022pattern} and virtual try-on~\cite{chong2021tryon,majithia2022wacv}.

In this paper, we focus on the problem of aligning garments. Existing methods for aligning garments are mostly extrinsic~\cite{bhatnagar2019iccv,ma2020cape,ponsmoll2017clothcap,tiwari2020sizer,xiang2022dressing,xiang2021separate,zhang2017clothed}. These methods directly operate in the extrinsic space, \ie, the 3D space, typically by fitting a template shape to target shapes in a non-rigid iterative closest point (ICP) manner. Extrinsic methods easily suffer from incorrect point matches, making it difficult to align details and even the overall shape. This problem becomes more severe for garments, due to their complex deformations as well as articulated motions. 

On the other hand, intrinsic methods~\cite{attaiki2021dpfm,donati2020dgfm,hamidian2020eigenreg,litany2017dfm,mateus2008lapeigen,ovsjanikov2012fm,rustamov2007lboeigen} leverage intrinsic manifold properties, most commonly the eigenfunctions of the Laplace-Beltrami operator (LBO)~\cite{meyer2003cotlaplacian}, which are independent of the extrinsic shape and therefore also free from incorrect matches in the extrinsic space. These high-dimensional intrinsic embeddings are also smoother than their extrinsic counterparts, and are easier to align. Current mainstream methods for this purpose are mostly based on the Functional Maps framework~\cite{ovsjanikov2012fm}, where one estimates a linear functional map between two intrinsic embeddings.
Unfortunately, directly applying the Functional Maps framework~\cite{ovsjanikov2012fm} for intrinsic garment alignment still poses challenges. Since a functional map typically needs to be computed using shape descriptor constraints~\cite{aubry2011wks,salti2014shot,sun2009hks}, they are more suitable for as-rigid-as-possible deformations. Garments, on the other hand, can stretch, shear and bend, rendering shape descriptors unreliable for functional map estimation. Moreover, the linear assumption of a functional map becomes insufficient for garments due to their non-isometric deformations.

In this work, we resolve these challenges by proposing two neural deformation fields to align the intrinsic embeddings of garments in a coarse-to-fine manner. Our method works in two stages. In the first stage, we use a neural deformation field to obtain a coarse 3D alignment of the source and the target, which induces a functional map to align their intrinsic embeddings. This bypasses the need for descriptors to estimate functional maps as in previous work~\cite{attaiki2021dpfm,donati2020dgfm,litany2017dfm,ovsjanikov2012fm}. Furthermore, we use an intrinsic neural field~\cite{koestler2022inf} to implement this deformation. Leveraging intrinsic features makes this deformation field robust to defects such as self-contact or self-intersections of the base template, allowing us to handle difficult garment types, \eg, long coats. In the second stage, the intrinsic alignment is further refined with a second neural deformation field. This differs from traditional (linear) functional maps by introducing non-linearity to the Functional Maps framework. We remark that the necessity of introducing non-linearity for garments is rooted in their highly non-rigid and non-isometric deformations. An analysis of the necessity of non-linearity will be provided in the supplementary materials. We demonstrate that our two-stage pipeline can align garment to the extent that both geometry and texture are accurately matched (Fig.~\ref{fig:teaser}). To summarize, our contributions include:

\begin{itemize}
    \item We propose a two-stage pipeline for aligning highly non-rigid 3D garment data, leveraging intrinsic manifold properties and neural deformation fields to achieve high-accuracy results.
    \item In the first stage, we propose an intrinsic neural deformation field that leverages intrinsic manifold properties. Such intrinsic fields do not suffer from self-contact or self-intersections of the base template. To the best of our knowledge, we are the first to use intrinsic neural fields for modeling deformations.
    \item In the second stage, we propose another neural deformation field that introduces non-linearity to the Functional Maps framework. This allows us to achieve texture-level high-accuracy alignment. We also provide a theoretical analysis why this is necessary in the supplementary material.
    \item We collect a dataset of high-quality 3D garment data, including difficult garment types such as long coats.
\end{itemize}

%% file: figs/fig1_tex.tex
\begin{figure}[t]
\begin{center}
   \includegraphics[width=\linewidth]{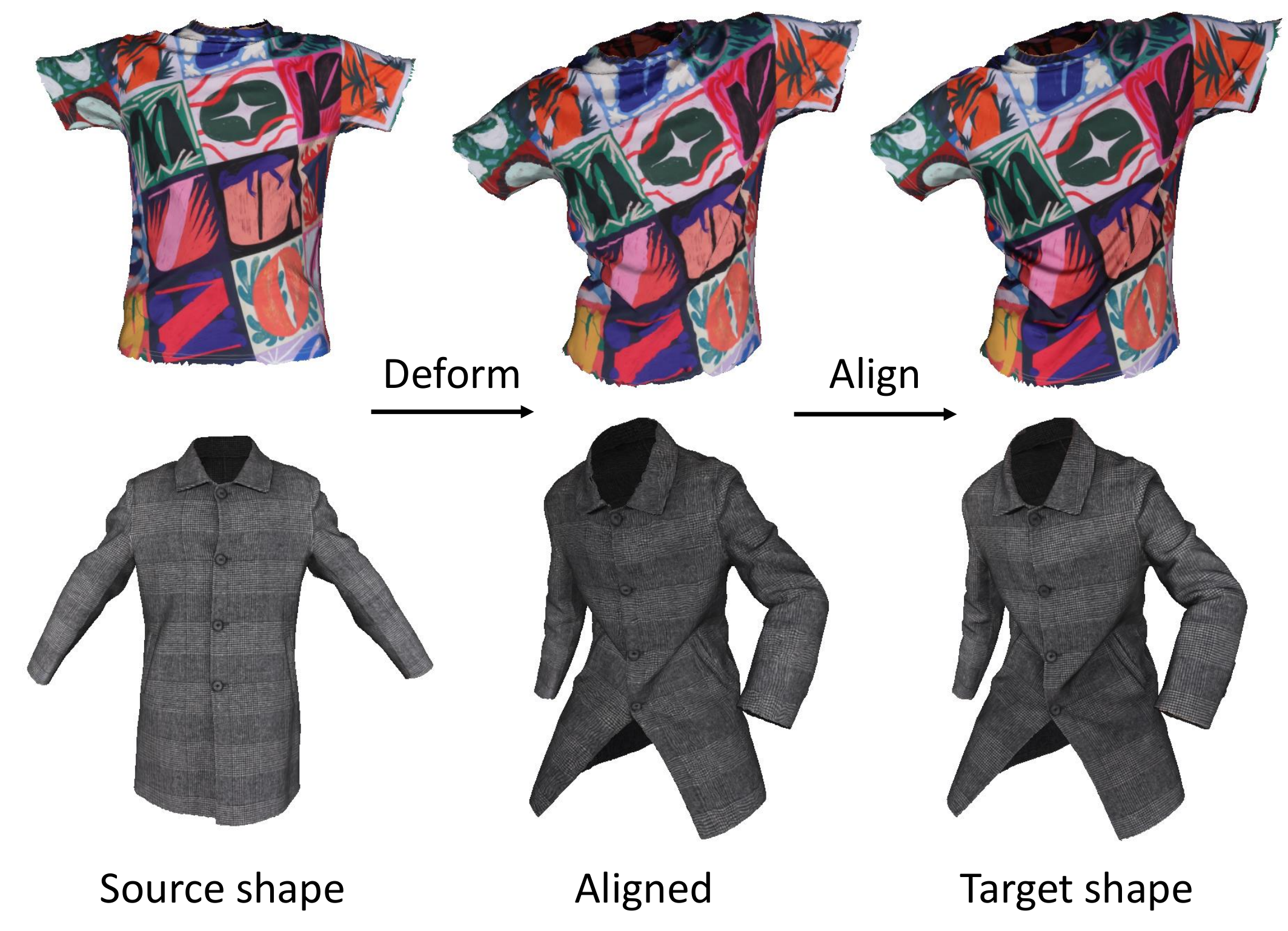}
\end{center}
   \caption{Our method can align deformed versions of a garment to the extent that both texture patterns and geometric details are accurately matched. Various types of garments can be handled, including difficult ones such as coats.}
\label{fig:teaser}
\end{figure}

%% file: sections/sec2_relate_v3.tex
\section{Related Work}

The problem of garment alignment falls into the larger category of non-rigid shape alignment/registration. We focus on some highly-related methods in this section. For a complete review we refer the readers to the survey of Deng \etal~\cite{deng2022nrregsurvey}.

\subsection{General Non-rigid Shape Alignment}

\textbf{Extrinsic methods} for shape alignment directly operate in the extrinsic space, \ie, the 3D space, typically by deforming a template to a target shape in a non-rigid ICP manner. Common choices for modeling deformations include direct vertex optimization (vertices as free variables) with different regularizations (differential coordinates, local affinity, as-rigid-as-possible, \etc)~\cite{amberg2007optimal,huang2011nonrigid,liao2009nonrigid,yamazaki2013nonrigid}, embedded deformations~\cite{bozic2021ndg,guo2015nonrigid,li2009nonrigid,sumner2007embedded,tretschk2020demea} and neural deformation fields~\cite{li2022ndp,yang2021gpnf}. 

Closely related to our work are neural deformations~\cite{li2022ndp,yang2021gpnf}, typically implemented as coordinate-based networks. They enjoy both flexibility and regularity due to being non-linear, continuous and having low-frequency bias~\cite{basri2019spectralbias,rahaman2019spectralbias}. However, coordinate-based neural deformations are continuous w.r.t. the ambient 3D space. If two parts are close in the 3D space, a neural deformation field can hardly pull them apart even if they are geodesically distant.
On observing this limitation of coordinate-based networks for other tasks, Koestler \etal~\cite{koestler2022inf} proposed intrinsic neural fields which are suited for representing fields on manifolds, and exhibited superior results in texture learning. 

\textbf{Intrinsic methods} utilize intrinsic embeddings, namely the embeddings derived from the eigenfunctions of the Laplace-Beltrami operator (LBO), to obtain shape correspondences and alignment~\cite{attaiki2021dpfm,donati2020dgfm,hamidian2020eigenreg,litany2017dfm,mateus2008lapeigen,ovsjanikov2012fm,rustamov2007lboeigen}. Intrinsic embeddings are independent of the extrinsic appearances of the underlying manifolds, and thus do not suffer from incorrect closest point matches faced by extrinsic methods. However, the LBO eigenfunctions of two near-isometric shapes may not be in one-to-one correspondences due to sign ambiguity, eigenvalue switching and multiplicity of eigenvalues~\cite{hamidian2020eigenreg,ovsjanikov2012fm,shi2013eigenmultiplicity}. 

The Functional Maps (FM) framework~\cite{ovsjanikov2012fm} solves for a linear transformation between the LBO eigenbases of two shapes. Functional map estimation typically requires shape descriptors~\cite{aubry2011wks,salti2014shot,sun2009hks}, whose quality is essential in the performance of functional maps. For this reason, a large body of existing work focuses on improving descriptors~\cite{attaiki2021dpfm,corman2015superviseddescriptor,donati2020dgfm,halimi2019dsc,litany2017dfm,roufosse2019ssm}. However, for highly non-rigid garments, shape descriptors become unreliable.
Aside from descriptors, the function map itself can also be refined for better accuracy. Ovsjanikov \etal~\cite{ovsjanikov2012fm} straightforwardly refine a functional map with linear ICP in the intrinsic space. Ren \etal~\cite{ren2018bcicp} improve this procedure by enforcing bijectivity and continuity. Spectral upsampling techniques can obtain functional correspondences in a coarse-to-fine manner~\cite{eisenberger2020smoothshells,eisenberger2020deepshells,melzi2019zoomout}. Even though intrinsic shape correspondences have been extensively studied, existing work focuses on near-isometric or as-rigid-as-possible deformations. To the best of our knowledge, there has been no intrinsic method that is devoted to high-accuracy garment alignment, which involves highly non-rigid wrinkle deformations.

\subsection{Human and Garment Shape Alignment}

There has been extensive research into human performance capture that reconstructs and tracks non-rigid clothed human surface geometries~\cite{guo2015nonrigid,newcombe2015dynamicfusion,slavcheva2017killingfusion,slavcheva2018sobolevfusion,su2020robustfusion,yu2018doublefusion}. These methods generally deal with full-body geometries and use tracking only to aid the reconstruction process. Another line of work~\cite{bhatnagar2020loopreg,ma2020cape,ma2021pop,ma2021scale} leverages the parametric body model SMPL~\cite{loper2015smpl} to represent clothing as an offset layer of naked body. While this introduces alignment in the sense that all garment deformations share a common body template, such an alignment does not track the actual movement of a point. Moreover, these methods do not segment body and garments, resulting in limited realism.

Despite the progress in reconstructing and animating full-body humans, recent work~\cite{halimi2022pattern,su2023caphy,xiang2022dressing,xiang2021separate} shows using aligned garment data allows human avatars to be presented with higher realism. ClothCap~\cite{ponsmoll2017clothcap} made an early attempt in this direction by aligning garments from 4D scans using a template segmented from the SMPL~\cite{loper2015smpl} model. SimulCap~\cite{yu2019simulcap} jointly captures the naked body and the deformed garment with multi-layered meshes to achieve plausible results for both body-cloth interaction and non-rigid tracking.

Unlike purely articulated shapes, garments contain detailed wrinkles where one faces an unavoidable trade-off between deformation flexibility and regularity. This difficulty is almost faced by all extrinsic methods. On the other hand, we are unaware of any intrinsic method that is devoted to garment alignment. Our work makes an attempt in this direction by leveraging intrinsic manifold properties for non-rigid garment alignment. 

%% file: sections/sec3_method.tex
\section{Method}

\input{figs/fig2_tex}

We seek to align deformed versions of a garment in different poses. We assume that garment data are presented as 3D triangular meshes with known SMPL~\cite{loper2015smpl} pose parameters $\theta\in\mathbb R^{72}$. Fig.~\ref{fig:pipeline} shows an overview of our pipeline. Let us denote the source shape by $M$ and the target shape by $N$. In the coarse stage, a smooth template is obtained from the source $M$, and is then used to align with the target $N$ in the 3D space (Sec.~\ref{sec:method-fminit}). This alignment induces a functional map~\cite{ovsjanikov2012fm}, which approximately aligns the intrinsic embeddings. In the refinement stage, we further refine the intrinsic alignment to obtain dense correspondences with a non-linear neural deformation field (Sec.~\ref{sec:method-align}). Finally, the vertex coordinates of $N$ are transferred to $M$ through the shape correspondences obtained from the intrinsic alignment (Sec.~\ref{sec:shapetransfer}). Details of network architecture and parameter selection are left to the supplementary material.

\subsection{Preliminary: Functional Maps}\label{sec:method-fm}
The Functional Maps framework~\cite{ovsjanikov2012fm} is a shape correspondence representation that has been extended significantly due to its compactness and flexibility~\cite{attaiki2021dpfm,corman2015superviseddescriptor,donati2020dgfm,eisenberger2020deepshells,eisenberger2020smoothshells,halimi2019dsc,litany2017dfm,melzi2019zoomout,ren2018bcicp,roufosse2019ssm}. We briefly review it to introduce terminology and fix notations. Interested readers can refer to \cite{ovsjanikov2012fm} for more details. Given two manifolds $M$ and $N$, together with a point-to-point map $P:M\to N$, for any function $g\in\mathcal F(N)$ (the function space on $N$), we can define a corresponding function on $M$ as $f:=g\circ P\in\mathcal F(M)$. Thus, any point-to-point map $P:M\to N$ between manifolds \emph{induces} a functional map $P_F:\mathcal F(N)\to\mathcal F(M),\ g\mapsto g\circ P$ between function spaces. Since any induced functional map must be linear as shown in \cite{ovsjanikov2012fm}, if two bases $\Phi^M$ and $\Phi^N$ are chosen for $\mathcal F(M)$ and $\mathcal F(N)$, respectively, then the functional map (and thus also the point-to-point map), can be represented as a (possibly infinite) matrix. In practice, $\Phi^M$ is often chosen to be the eigenfunctions of the Laplace-Beltrami operator (LBO)~\cite{meyer2003cotlaplacian}.

In the discrete case, let $M$ be represented by a vertex matrix $V^M\in{\mathbb R}^{n_V\times3}$ and a face matrix $F^M\in{\mathbb N}^{n_F\times3}$. A function $f\in\mathcal F(M)$ is then a scalar array $f\in\mathbb R^{n_V}$. The LBO is then a matrix $L\in\mathbb R^{n_V\times n_V}$. We use cotangent weighting~\cite{meyer2003cotlaplacian} to discretize the LBO as $\mathcal L=A^{-1}W$, where $A_{ii}$ is the Voronoi area near vertex $i$ and
\begin{equation}
W_{ij}=\left\{
\begin{array}{lll}
-\frac{\cot\alpha_{ij}+\cot\beta_{ij}}{2}&(i,j)\textrm{ is an edge}&\\
\sum_{(i,k)\textrm{is an edge}}\frac{\cot\alpha_{ik}+\cot\beta_{jk}}{2}&i=j&\\
0&\textrm{otherwise}.&
\end{array}
\right.
\end{equation}
Here, $\alpha_{ij}$ and $\beta_{ij}$ are the angles opposite to the edge $(i,j)$.
It is well known that eigenvalues of $\mathcal L$ are non-negative, with a natural ordering given by: $\lambda_0=0<\lambda_1\leq\lambda_2\leq...$, and that the eigenvectors $\{\phi_j\}_{j\geq0}$ form an orthonormal (w.r.t. the $A$-inner product) basis of $\mathcal F(M)$.
Let
\begin{equation}
\Phi^{M}=[\phi_1,\phi_2,\cdots,\phi_K]\in\mathbb R^{n_V\times K}\label{eq:phim}
\end{equation}
denote the first $K$ non-constant eigenfunctions on $M$ stacked together (omitting $K$ for simplicity). Then the columns of $\Phi^{M}$ spans a subspace of ${\rm span}(\Phi^{M})$ of $\mathcal F(M)$. Let these notations be defined for $N$ similarly, then any functional map ${\rm span}(\Phi^{N})\to{\rm span}(\Phi^{M})$ can be represented as a matrix $C\in\mathbb R^{K\times K}$.

\subsection{Coarse Alignment in the 3D Space}\label{sec:method-fminit}

In this section, we obtain a coarse alignment of the source and the target in the 3D space. The closest point correspondences from this alignment will be used to obtain a function map for aligning intrinsic embeddings. We start from obtaining a rigged smooth template and then use this template as a substitute of $M$ to align with the chosen target mesh (coarse stage in Fig.~\ref{fig:pipeline}).

\paragraph{Rigged smooth template acquisition}

Following ClothCap~\cite{ponsmoll2017clothcap}, to avoid any potential negative effect of pose-dependent wrinkles in the source mesh, prior to 3D fitting we obtain a smoothed template rigged to the SMPL~\cite{loper2015smpl} skeleton from an approximately A-pose scan. Let ${\rm LBS}(v,w,\theta)$ denote the linear blend skinning (LBS) transformation as in \cite{loper2015smpl}, \ie, it moves a point $v\in\mathbb R^3$ in the T-pose space to the pose space (defined by the SMPL~\cite{loper2015smpl} pose parameter $\theta$), with $w\in\mathbb R^{24}$ being its skinning weights.
Let $M=(V^M,F^M)$ denote the source mesh and $\theta^M$ its corresponding SMPL pose parameters. The desired smooth template is another mesh $T=(V^T,F^M)$ having the same topology as $M$, with skinning weights $W$, s.t. when it is posed to $\theta$, it aligns with $M$. More specifically, we let $V^T$ minimize the following energy:
\begin{eqnarray}
E_1(V^T)&=&w_1\|{\rm LBS}(V^T,W(V^T),\theta^M)-V^M\|_F^2\nonumber\\
&&+w_2\|E^T-E^M\|_F^2+w_3\|L^MV^T\|_F^2,
\end{eqnarray}
where $w_i$ are weights to balance different energy terms. Here, we use the diffused skinning field $W(\cdot)$ in \cite{lin2022fite} to ensure smoothness. $E^T$ and $E^M$ are arrays of edge lengths, which can be computed from $(V^T,F^M)$ and $(V^M,F^M)$. Note that the edges of $T$ and $M$ are in one-to-one correspondence since they have identical face lists. $L^M$ is the uniform Laplacian of $M$. The first term forces the posed vertices of $T$ to be close to $M$. The second term enforces edge regularization for $M$ and $T$ to be isometric. The last term enforces smoothness.

\paragraph{Coarse fitting with neural deformation field}
Having obtained the rigged smooth template $T$, which has identical topology to that of the source garment mesh $M$, we use it as a smooth substitute for $M$ to align with any other target garment mesh $N$. Let $\theta^N$ denote the body pose corresponding to $N$.
We first pose $T$ with $\theta^N$ to obtain $T'=(V',F^M)$, where
\begin{equation}
    V'=LBS(V^T,W(V^T),\theta^N)
\end{equation}
We then fit $T'$ to $N$ by adding a neural deformation field to it.
Note that this deformation field only needs to be defined on the manifold $M$, not the entire 3D space. We thus let this deformation to be modeled as an intrinsic neural field~\cite{koestler2022inf} $D_\tau$ (see Fig.~\ref{fig:pipeline}). Leveraging intrinisic features, \ie, the LBO eigenfunctions, allows us to define a manifold deformation field independent of how it is embedded in the 3D space. More specifically, an intrinsic deformation field does not suffer from self-contact or self-intersection of the posed template $T'$ and allows more flexible deformations. The vertices $V'$ of the posed template $T'$ are further deformed to
\begin{equation}
    V''=V'+D_\tau(\Phi^{M}),\quad D_\tau:\mathbb R^K\to\mathbb R^3.
\end{equation}
Note that $n_V$ is considered as a batch dimension for $\Phi^{M}$ (see Eq.~\eqref{eq:phim}) when writing $D_\tau(\Phi^{M})$.\footnote{Throughout the paper we use the convention that if $f(\cdot):\mathbb R^l\to\mathbb R^q$ is a mapping defined on $\mathbb R^l$ and $V\in\mathbb R^{n\times l}$ is a matrix, then we write $f(V)$ to mean that $f$ applies to each row of $V$ independently, with the results again stacked together as a matrix, \ie, $f(V)\in\mathbb R^{n\times q}$.}
Here, $\tau$ denotes the optimizable parameters of $D_\tau$. Let $V''_{\rm b}$ denote the subset of $V''$ containing only the boundary vertices (same for $V^M_{\rm b}$). We minimize the following energy:
\begin{eqnarray}
E_2(\tau)&=&w_4{\rm CD}(V'',V^N)+w_5{\rm CD}(V''_{\rm b},V^N_{\rm b})\nonumber\\
&&+w_6\|\max(E^{T''}-E^M,0)\|_F^2,\label{eq:loss-2}    
\end{eqnarray}
where ${\rm CD}(\cdot,\cdot)$ denotes the Chamfer distance between point clouds and $E^{T''}$ denotes the edge lengths of the posed and deformed template $T''=(V'',F^M)$. Note that we clip $E^{T''}-E^M$ by $0$ to allow squeezing, since the target shape $N$ may have invisible parts due to occlusion during capture.
After $\tau$ has been optimized, the deformed template $T''$ is coarsely fitted to $N$. This concludes the coarse fitting stage.

\input{figs/tbl1-average}

\subsection{Refinement in the Intrinsic Space}\label{sec:method-align}

In this section, we seek to accurately align the intrinsic embeddings of $M$ and $N$. Note that the initially computed intrinsic embeddings can differ greatly due to sign ambiguity and eigenvalue switching~\cite{hamidian2020eigenreg,ovsjanikov2012fm,shi2013eigenmultiplicity}. Intuitively, these can lead to the intrinsic embedding of $N$ in Fig.~\ref{fig:pipeline} appearing ``upside-down''. The coarse 3D correspondences from the last stage are thus used to induce a linear functional map $A_0$ (bottom-right corner of Fig.~\ref{fig:pipeline}), bringing two intrinsic embeddings into an approximate alignment. However, limited by its linearity, $A_0$ cannot achieve enough accuracy when the ground truth deformation is non-rigid and non-isometric. We thus use a second neural network (introducing non-linearity) to further refine the intrinsic alignment.

\paragraph{Functional map from coarse 3D correspondences}

We obtain the functional map $A_0$ as follows. First, we search for each vertex $i$ in $V''$ its nearest neighbor $j(i)$ in $V^N$, giving rise to a point-to-point map $i\mapsto j(i)$. Then we reindex $\Phi^{N}$ to obtain $\hat\Phi^{M}\in\mathbb R^{n_V\times K}$ as follows:
\begin{equation}
    \hat\Phi^{M}_{ik}:=\Phi^{N}_{j(i),k}
\end{equation}
As defined in \cite{ovsjanikov2012fm}, the functional map $A_0\in\mathbb R^{K\times K}$ induced by $i\mapsto j(i)$ satisfies
\begin{equation}
    \hat\Phi^{M}A_0=\Phi^{M}.
\end{equation}
In practice this is solved in a least square sense. We then apply $A_0$ to rectify the intrinsic embedding of $N$ as:
\begin{equation}
    \hat\Phi^{N}=\Phi^{N}A_0.\label{eq:init-A0}
\end{equation}

\paragraph{Refining the intrinsic alignment}
The initial functional map $A_0$ in \eqref{eq:init-A0} can bring the intrinsic embeddings of $M$ and $N$ to an approximate alignment, which we further refine with a second neural deformation network (the refining stage in Fig.~\ref{fig:pipeline}).
Unlike previous work~\cite{ovsjanikov2012fm} that estimates a linear transformation between $\Phi^{M}$ and $\hat\Phi^{N}$, our neural deformation field in the intrinsic space introduces non-linearity. We remark that it is necessary to introduce non-linearity if high accuracy is desired. A detailed analysis is given in the supplementary material.

Prior to refining the alignment, we follow \cite{rustamov2007lboeigen} to scale down $\phi_j$ to $\phi_j/\sqrt{\lambda_j}$, where $\lambda_j$ is its corresponding eigenvalue. This is because high-frequency components (corresponding to large eigenvalues) may lead to incorrect closest point matches when computing the Chamfer distances in Eq.~\eqref{eq:loss-3}. We use $\Phi^{M\downarrow}$ denote the eigenfunctions that have been scaled down (similarly defined for $\Phi^N$, $\hat\Phi^{N}$, \etc). Then we deform $\Phi^{M\downarrow}$ using a neural network $D_\varphi$:
\begin{equation}
\hat\Phi^{M\downarrow}=\Phi^{M\downarrow}+D_\varphi(\Phi^{M\downarrow}).
\end{equation}
Here, $\varphi$ denotes the parameters of $D_\varphi$, which are optimized to minimize:
\begin{eqnarray}
E_3(\varphi)&=&w_7{\rm CD}(\hat\Phi^{M\downarrow},\hat\Phi^{N\downarrow})+w_8{\rm CD}(\hat\Phi^{M\downarrow}_{\rm b},\hat\Phi^{N\downarrow}_{\rm b})\nonumber\\
&&+w_9\|D_\varphi(\Phi^{M\downarrow})\|^2_F.\label{eq:loss-3}
\end{eqnarray}
The first two terms are data terms for the whole shape and for boundaries. Here, CD denotes the obvious extension of Chamfer distance to high-dimensional points, \ie, the averaged squared $L_2$-distance from each point in one point cloud to its closest point in the other point cloud, computed in both directions. The third term regularizes the magnitude of deformation. Having optimized $\varphi$ for \eqref{eq:loss-3}, we obtain accurately aligned intrinsic embeddings $\hat\Phi^{M\downarrow}$ and $\hat\Phi^{N\downarrow}$, from which dense correspondences can be extracted via nearest-neighbor search in the intrinsic space.

\subsection{Shape Transfer}\label{sec:shapetransfer}

For garment alignment, a triangle mesh having the same topology as $M$ but aligned to $N$ is generally more desired.
A naive shape transfer method is to first find the shape correspondences using their aligned intrinsic embeddings $\hat\Phi^{M\downarrow}$ and $\hat\Phi^{N\downarrow}$ from the last step, and then transfer the 3D vertex coordinates of $N$ through these correspondences. However, we find this leads to noisy alignment. To reduce noise in the shape transfer process, we utilize 3D coordinates together with Laplacian coordinates. 

Leveraging the aligned intrinsic embeddings, we search for each point $i$ in $\hat\Phi^{M\downarrow}$ its $k$ nearest neighbors $k{\rm NN}(i)$ in $\hat\Phi^{N\downarrow}$, with $j(i)$ being the nearest one. Then we define
\begin{equation}
    \hat V^N_i=\frac{c_{ik}w_{ik}}{\sum_{k\in k{\rm NN}(i)}c_{ik}w_{ik}}V^N_k,
\end{equation}
where $w_{ik}$ is inversely proportional to the $L_2$ distance between $\hat\Phi^{M\downarrow}_i$ and $\hat\Phi^{N\downarrow}_k$, and $c_{ik}$ is set to a larger value if $i$ or $k$ corresponds to a boundary vertex. We let the desired vertex coordinates $\hat V^M$ for the alignment from $M$ to $N$ minimize: 
\begin{eqnarray}
    E_4(\hat V^M)&=&\sum_{i}\|\hat V^M_i-\hat V^N_i\|^2\nonumber\\
    &+&\sum_{i}\|(L^M\hat V^M)_i-(L^NV^N)_{j(i)}\|^2,
\end{eqnarray}
where $L^M$ and $L^N$ are the uniform Laplacian matrices on $M$ and $N$. In short, we transfer both 3D coordinates and Laplacian coordinates, and solve for the vertex positions that best satisfy both. Finally, the aligned mesh is obtained as $\hat M=(\hat V^M,F^M)$, which has the same topology as $M$, but aligns with $N$ in geometry.

\subsection{Texture Alignment}\label{sec:texture-align}
Since the deformations in both stages are neural networks, which enjoy continuity and low-frequency bias~\cite{basri2019spectralbias,rahaman2019spectralbias}, the deformations are automatically well-regularized. Texture alignment with neural deformations can easily be achieved by directly concatenating per-vertex RGB colors $C^M\in\mathbb R^{n_V\times3}$ to $\Phi^{M}$:
\begin{equation}
\widetilde\Phi^{M}=[\beta_1\Phi^{M},\beta_2C^M]\in\mathbb R^{n_V\times(K+3)}.
\end{equation}
where $\beta_1$ and $\beta_2$ are parameters to balance geometry and color information. We only concatenate colors when computing the first term in \eqref{eq:loss-2} and in \eqref{eq:loss-3}. Colors are not used as input to $D_\tau$ and $D_\varphi$, nor are they used in the boundary terms in \eqref{eq:loss-2} and \eqref{eq:loss-3}.

%% file: figs/fig2_tex.tex
\begin{figure*}[t]
\begin{center}
   \includegraphics[width=\linewidth]{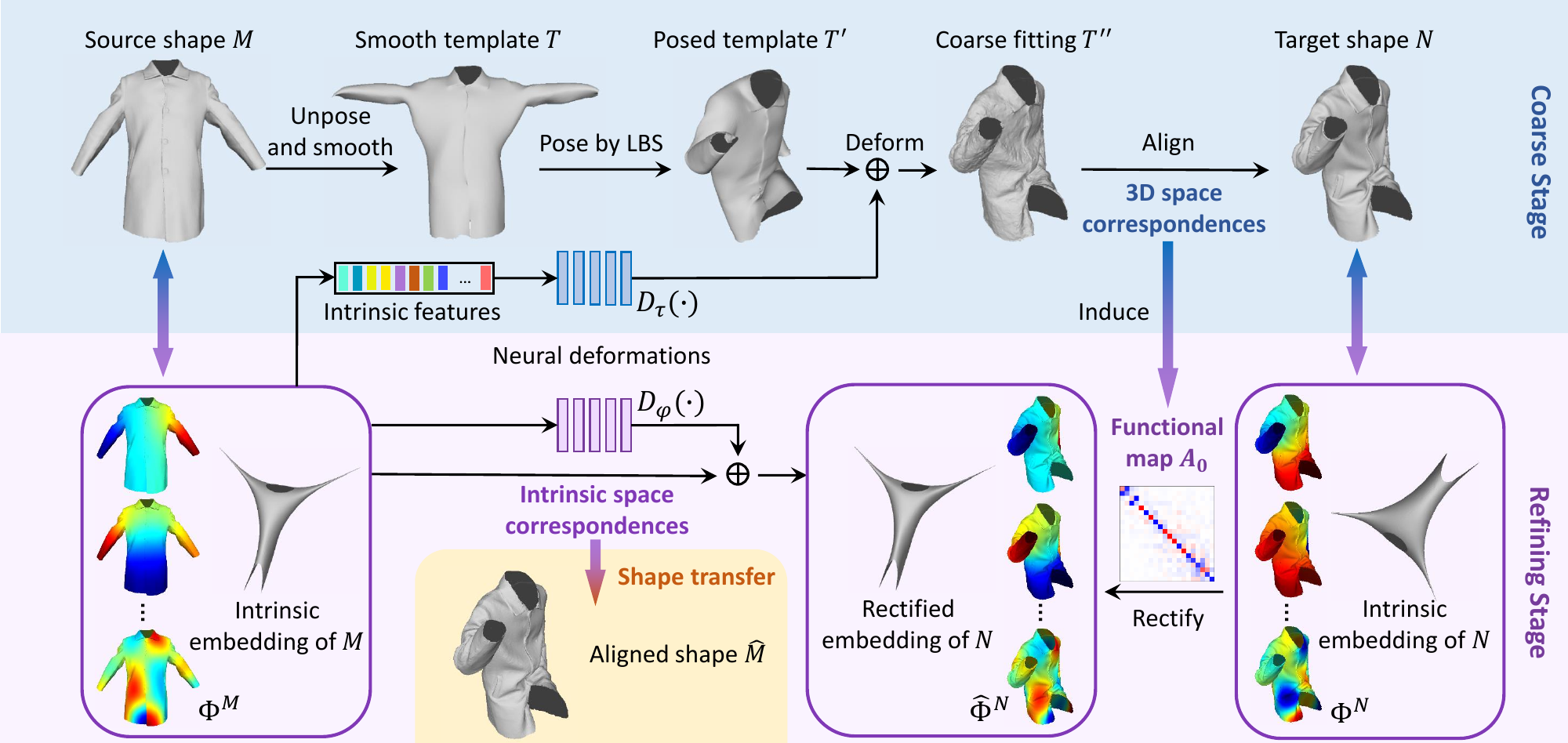}
\end{center}
   \caption{An overview of our pipeline. Given the source $M$ and the target $N$, we start from the 3D space where a neural deformation field with intrinsic input features deforms the template to align with the target (\textcolor{blue}{coarse stage}, Section~\ref{sec:method-fminit}). This coarse alignment induces a functional map that can bring the intrinsic embeddings of $M$ and $N$ into approximate alignment, which is further refined with a second neural deformation field (\textcolor{violet}{refining stage}, Section~\ref{sec:method-align}). Finally, we transfer the 3D vertex coordinates from $N$ to $M$ through the intrinsic space correspondences (\textcolor{brown}{shape transfer}, Section~\ref{sec:shapetransfer}).}
\label{fig:pipeline}
\end{figure*}

%% file: figs/tbl1-average.tex
\begin{table*}
\begin{center}
\begin{tabular}{|l|c|c|c|c|c|}
\hline
\multirow{2}*{Method} & Chamfer Dist. & Cos. Sim. & LPIPS-AlexNet & LPIPS-VGG & SSIM \\
 & ($\times10^{-3}$) $\downarrow$ & $\uparrow$ & ($\times10^{-2}$) $\downarrow$ & ($\times10^{-2}$) $\downarrow$ & $\uparrow$\\
\hline
NDP~\cite{li2022ndp}                 & 14.305 & 0.885 & 11.414 & 11.690 & 0.865 \\
ClothCap~\cite{ponsmoll2017clothcap} & 3.144  & 0.965 & 7.874  & 9.468  & 0.888 \\
Deep Shells~\cite{eisenberger2020deepshells} & 4.094  & 0.955 & 8.471  & 9.796  & 0.887 \\
Ours (NoTex)& \textbf{2.613} & \textbf{0.979} & \underline{5.053} & \underline{7.409} & \underline{0.901} \\
Ours        & \underline{2.627} & \textbf{0.979} & \textbf{4.318} & \textbf{6.616} & \textbf{0.914} \\
\hline
\end{tabular}
\end{center}
\caption{Quantitative comparison results. ``NoTex'' means not using texture information as done in Sec.~\ref{sec:texture-align}. Our method (with or without using texture information) performs notably better than baselines. The best results are in boldface and the second best results are underlined.\label{tbl:comparisons}}
\end{table*}

%% file: sections/sec4_exp.tex
\section{Experiments}

\input{figs/fig4_tex}

\subsection{Dataset and Baselines}

We introduce a new dataset, \textit{GarmCap}, containing high-quality textured 3D garment scans in various poses. This dataset includes four different garments (Fig.~\ref{fig:comparison}): G01 (a T-shirt with rich color patterns, 173 scans), G02 (a long coat with black-white strip patterns, 103 scans), G03 (a thick coat with a fading graywhite texture, 101 scans) and G04 (an orange coat, 119 scans). Data in \textit{GarmCap} are garments in static poses, collected in a cage with 128 cameras.

We evaluate our method with the \textit{GarmCap} dataset, and compare with ClothCap~\cite{ponsmoll2017clothcap}, Deep Shells~\cite{eisenberger2020deepshells} and Neural Deformation Pyramid (NDP)~\cite{li2022ndp}. Deep Shells~\cite{eisenberger2020deepshells} is the most related to our work since it also leverage both 3D-space and intrinsic-space alignments, but in a product space formulation. Also bearing similarity to ours, NDP~\cite{li2022ndp} models neural deformations in a hierarchical manner to achieve coarse-to-fine alignment. Please refer to the supplementary material for more details of the experimental setup.

\input{figs/fig6-7_tex}

\subsection{Comparisons}

We compare our method with the baselines introduced above. Since the \textit{GarmCap} dataset has no ground truth correspondences, we use Chamfer distance and cosine normal similarity to measure geometric similarity. For evaluating correspondence, we directly apply the texture of the source mesh to the aligned mesh and render 32 views per scan, and measure the similarity between rendered images and ground truth renderings. We use perceptual metrics (LPIPS)~\cite{zhang2018lpips} and structural similarity index measure (SSIM)~\cite{wang2003ssim}.

In Table~\ref{tbl:comparisons}, our method outperforms others in both geometry and texture alignments. Since the existing baseline implementations do not support using colors, we also test our method without using texture information as in Sec.~\ref{sec:texture-align} (labeled ``NoTex'' in Table~\ref{tbl:comparisons}). While doing so leads to a slight performance drop in texture alignment (see the LPIPS and SSIM metrics), there is no performance drop in geometric accuracy. Moreover, even without texture information, our method still surpasses all baseline methods. Fig.~\ref{fig:comparison} shows qualitative results of the alignments. While ClothCap~\cite{ponsmoll2017clothcap} and Deep Shells~\cite{eisenberger2020deepshells} are able to produce geometric details in the alignment, they may fail due to incorrect point matches and lead to visible defects, \eg, indentations on wrinkles (G01), wrongly aligned garment parts (G02), unnatural distortions (G03 and G04), and a number of inverted or self-intersecting faces. Deep Shells also occasionally suffer from symmetry issues (the left arm and the right arm are switched in G03). This is possibly due to its reliance on SHOT descriptors~\cite{salti2014shot}, which does not disambiguate the left-right symmetry of garments. The alignment results of NDP~\cite{li2022ndp} does not produce much distortion. However, being too rigid also forbids wrinkles and other details. Our method not only recovers accurate geometric details, but also achieves texture-level alignment. This can be observed from S02: our texture of pocket is accurately overlaid over its corresponding geometry, while for other methods texture shift and distortion distortion are evident.

\subsection{Ablation Studies}

We conduct ablation studies to evaluate how certain modules affect the overall alignment quality. Note that the performance of each module can vary with garment types, \eg, the coarse-stage intrinsic deformation field handles difficult garments (S02) while texture alignment is more useful for rich-texture garments (S01). We thus report the metrics for each garment separately. We present some metrics for S02 in Table~\ref{tbl:abl-separate} since it is the most difficult case in our dataset that can clearly reflect the effect of different modules. We leave the complete results to the supplementary material due to space limits.

\paragraph{Different deformation models in stage one}
We replace the intrinsic deformation field in the coarse stage (stage 1) by direct vertex optimization (DVO-S1), \ie, optimize vertex coordinates as free variables. Note that DVO is essentially the same as ClothCap~\cite{ponsmoll2017clothcap}, but with our energy function~\eqref{eq:loss-2}. We also experiment with coordinate-based neural networks in stage 1 (CoordNet-S1). The first row in Fig.~\ref{fig:abl} shows that both directly optimizing vertices and using coordinate-based networks can introduce unwanted texture distortions. Compared with using a coordinate-based network in the coarse stage, our intrinsic network is better at handling more complicated garment types, namely, long coats (Table~\ref{tbl:abl-separate}). Another benefit of using intrinsic deformation networks is the robustness to poor initializations. As shown in Fig.~\ref{fig:wrongsmpl}, an inaccurate SMPL estimation can lead to a poorly initialized template $T'$, which notably differs from the target shape and even has self-intersections. Since intrinsic features enjoy geodesic continuity on $T'$, the intrinsic deformation network allows us to pull away the self-intersecting parts and still obtain good results. Due to being robust to inaccurate SMPL estimations and template initializations, our method can also be applied loose garments where SMPL-driven the LBS initialization is not fully compatible (Fig.~\ref{fig:dress}).

\input{figs/fig9-rebuttal-smpl_tex}
\input{figs/fig10-rebuttal-dress_tex}

\paragraph{Different refinement methods in stage two}
We replace our non-linear neural refinement procedure in the refining stage (stage 2) by linear ICP refinement~\cite{ovsjanikov2012fm} (denoted as LR-S2, for ``linear refinement in stage 2''), and no refinement at all (NR-S2, for ``no refinement in stage 2''). The second row in Fig.~\ref{fig:abl} shows that our non-linear refinement can lead to a more uniform mesh triangulation in the final alignment than other methods. Quantitative results in Table~\ref{tbl:abl-separate} show that our non-linear refining strategy can improve both geometric and rendering results.

\paragraph{Texture information} We also test our pipeline without using texture information in Section~\ref{sec:texture-align}, labeled NoTex in Table.~\ref{tbl:abl-separate}. As shown in Table~\ref{tbl:abl-separate}, while this does not affect geometric accuracy, texture alignment becomes less accurate.

\input{figs/tbl2-singlecol}

%% file: figs/fig4_tex.tex
\setlength{\unitlength}{\textwidth}
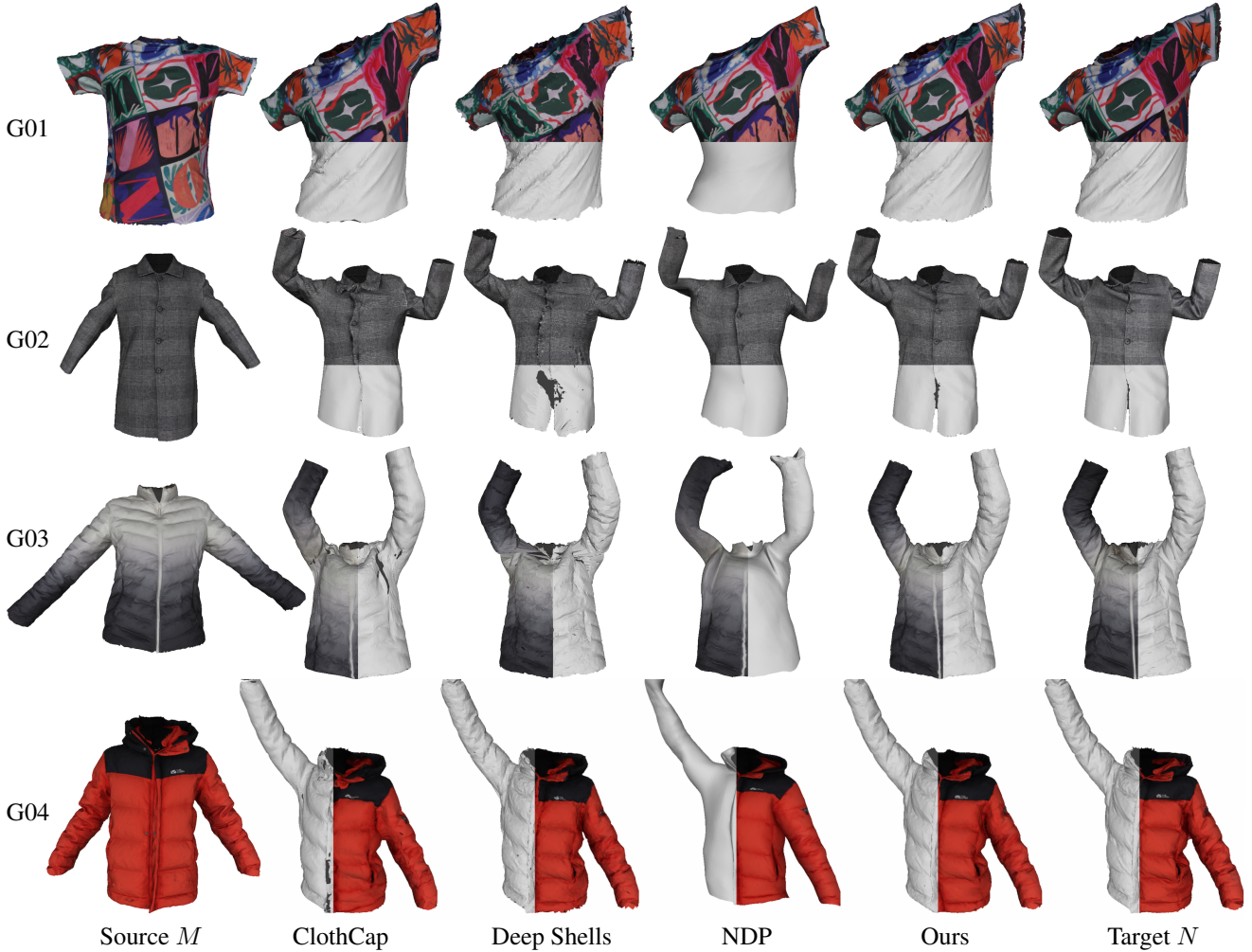
\begin{figure*}[t]
  \begin{center}
  \begin{picture}(1,0.763)(0,0)
  \put(0,0.02){\includegraphics[width=\linewidth]{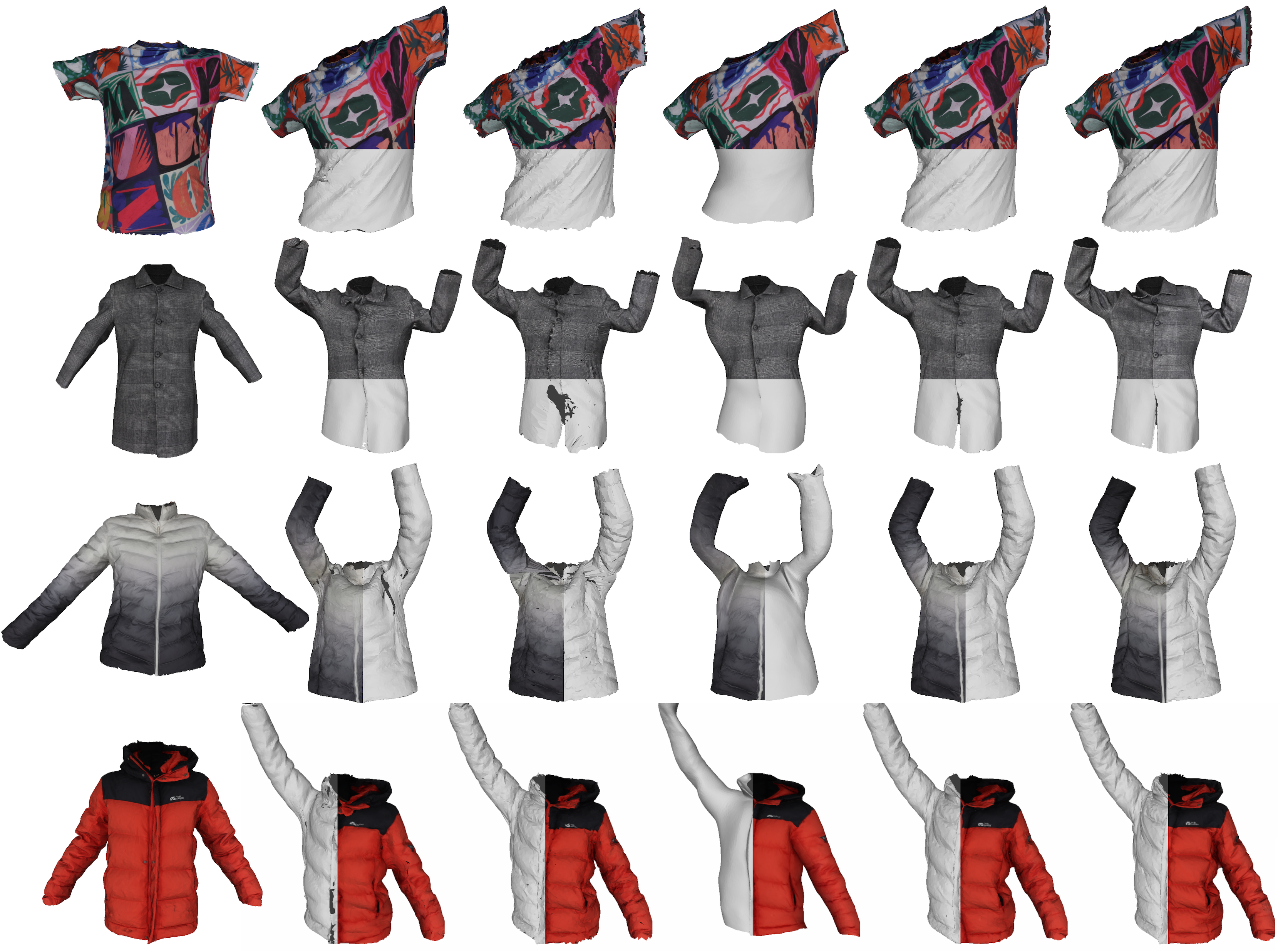}}
  \put(0.0,0.65){G01}
  \put(0.0,0.48){G02}
  \put(0.0,0.32){G03}
  \put(0.0,0.1){G04}
  \put(0.075,0.0){Source $M$}
  \put(0.23,0.0){ClothCap}
  \put(0.39,0.0){Deep Shells}
  \put(0.575,0.0){NDP}
  \put(0.735,0.0){Ours}
  \put(0.885,0.0){Target $N$}
  \end{picture}
  \end{center}
\caption{Qualitative comparisons. We exhibit both texture and geometry results. ClothCap~\cite{ponsmoll2017clothcap} is too flexible and produces unnatural distortions and inverted faces. Deep Shells~\cite{eisenberger2020deepshells} sometimes infers incorrect correspondences, possibly due to its reliance on shape descriptors. NDP~\cite{li2022ndp} is too rigid to produce wrinkles. Our method achieves both geometric and texture alignment.}
\label{fig:comparison}
\end{figure*}

%% file: figs/fig6-7_tex.tex
\setlength{\unitlength}{\textwidth}
\begin{figure*}[t]
\begin{center}
  \begin{picture}(1,0.475)(0,0)
  \put(0,0.02){\includegraphics[width=\linewidth]{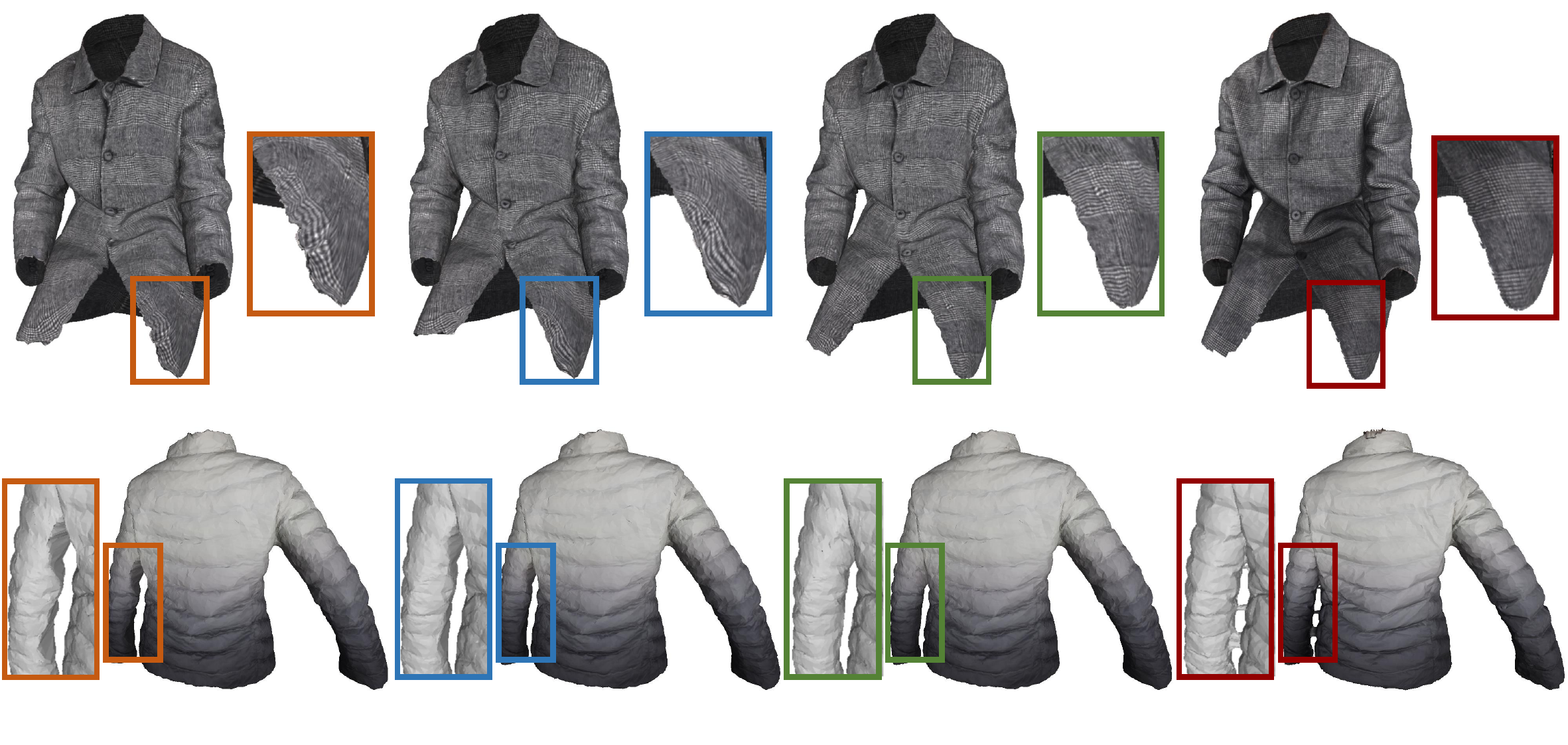}}
  \put(0.08,0.23){DVO-S1}
  \put(0.3,0.23){CoordNet-S1}
  \put(0.6,0.23){Ours}
  \put(0.85,0.23){Target $N$}
  \put(0.083,0.03){NR-S2}
  \put(0.33,0.03){LR-S2}
  \put(0.6,0.03){Ours}
  \put(0.85,0.03){Target $N$}
  \end{picture}
\end{center}
   \caption{Qualitative ablation studies. First row: using different deformation models for stage 1. Second row: using different refinement method for stage 2.}
\label{fig:abl}
\end{figure*}
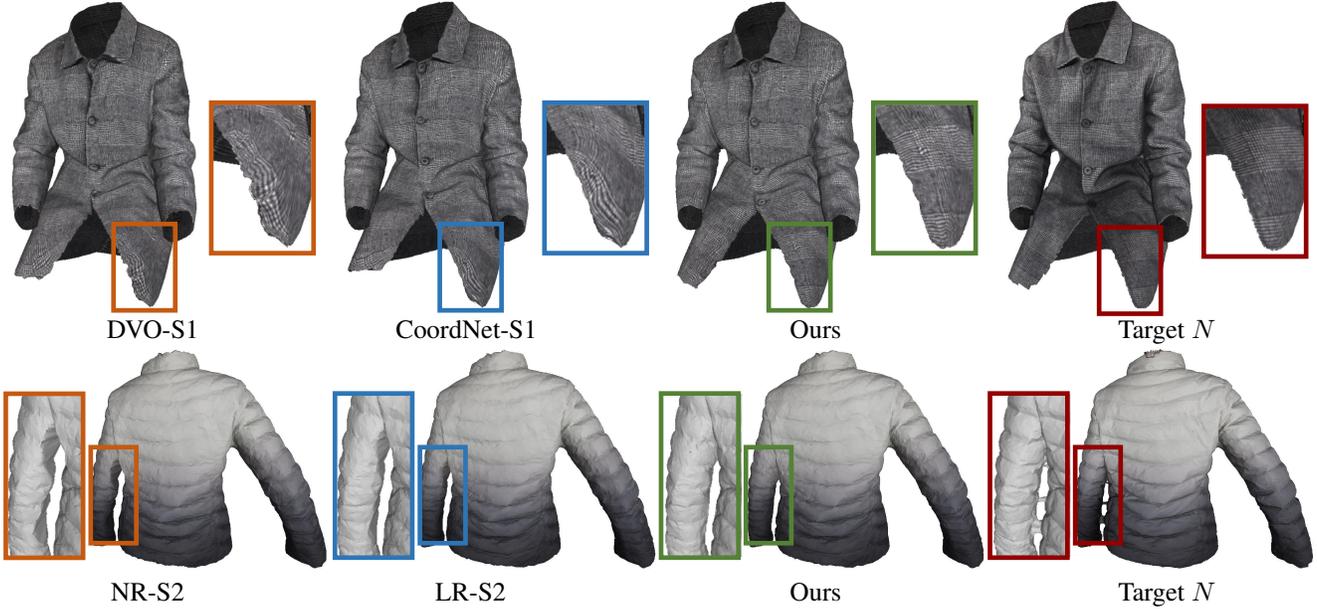

%% file: figs/fig9-rebuttal-smpl_tex.tex
\setlength{\unitlength}{\linewidth}
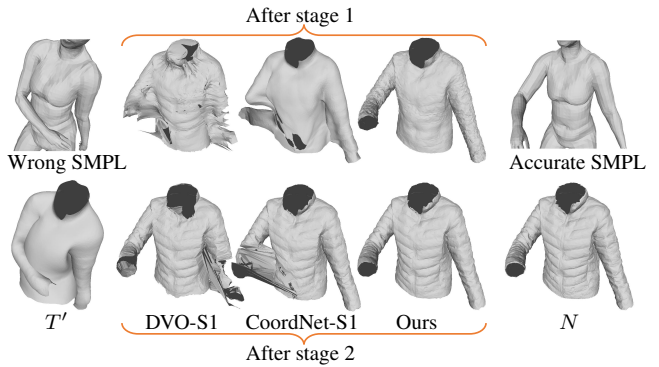
\begin{figure}
  \begin{center}
  \begin{picture}(1,0.545)(0,0)
  \put(0,0.025){\includegraphics[width=\linewidth]{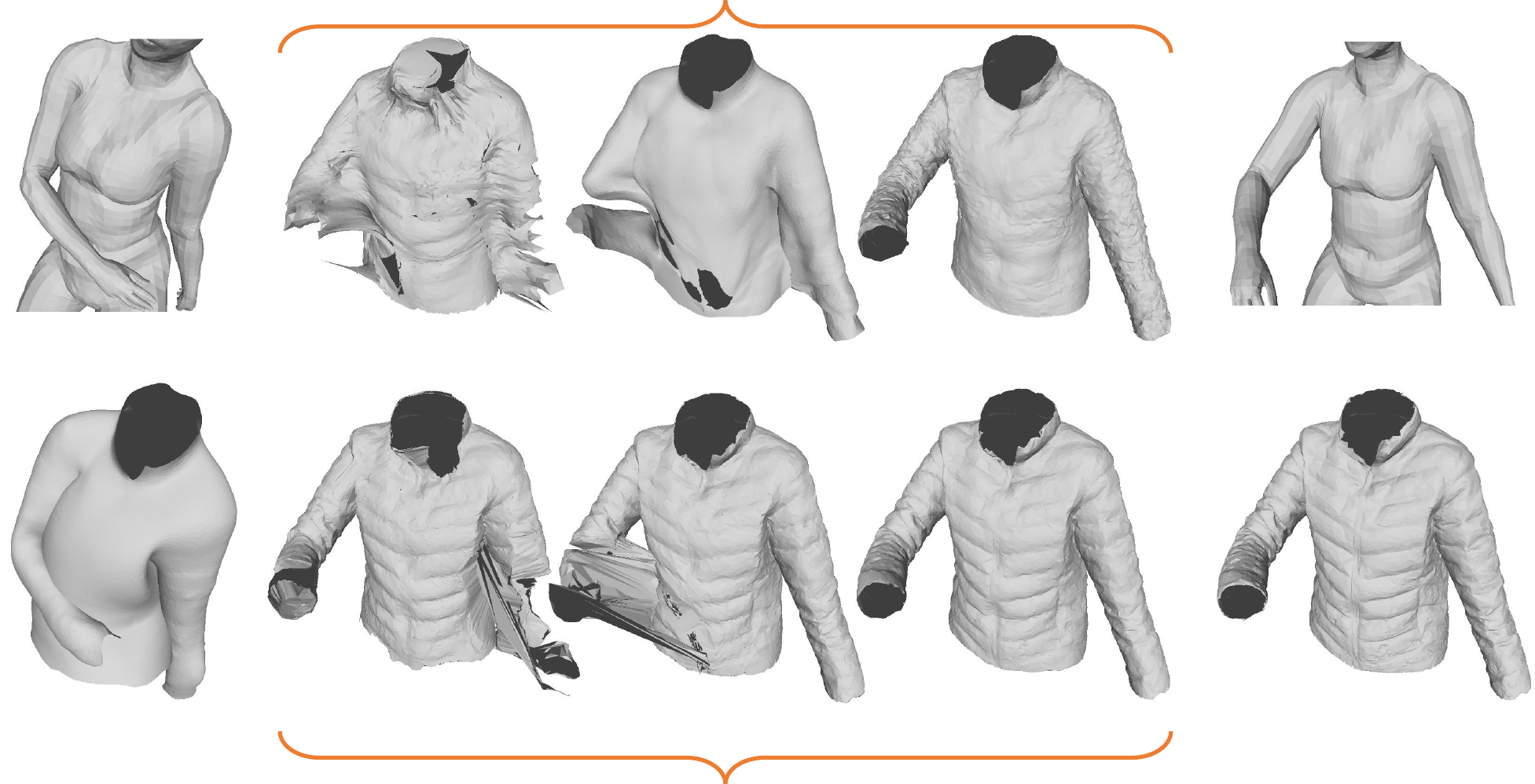}}
  \put(0.38,0.54){\footnotesize After stage 1}
  \put(0.38,0.0){\footnotesize After stage 2}
  \put(0.0,0.305){\footnotesize Wrong SMPL}
  \put(0.06,0.05){\footnotesize $T'$}
  \put(0.22,0.05){\footnotesize DVO-S1}
  \put(0.38,0.05){\footnotesize CoordNet-S1}
  \put(0.62,0.05){\footnotesize Ours}
  \put(0.8,0.305){\footnotesize Accurate SMPL}
  \put(0.88,0.05){\footnotesize $N$}
  \end{picture}
  \end{center}
\caption{Robustness to inaccurate SMPL estimations. Leftmost column: the template $T'$ initialized by LBS with an inaccurate SMPL. Middle columns: alignment results after stage 1 and stage 2 for different deformation modules used in stage 1. Rightmost column: The target shape $N$ and its corresponding accurate SMPL estimation.}
\label{fig:wrongsmpl}
\end{figure}

%% file: figs/fig10-rebuttal-dress_tex.tex
\setlength{\unitlength}{\linewidth}
\begin{figure}
  \begin{center}
  \begin{picture}(1,0.37)(0,0)
  \put(0,0.04){\includegraphics[width=\linewidth]{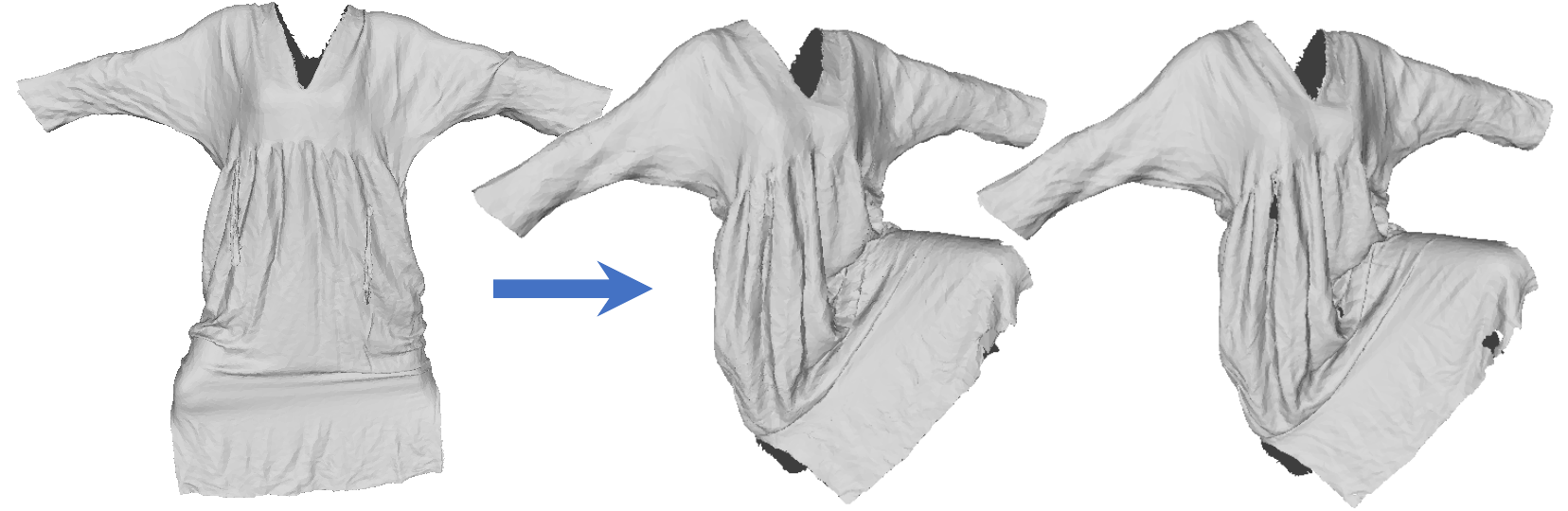}}
  \put(0.1,0.0){Source $M$}
  \put(0.4,0.0){Alignment $\hat M$}
  \put(0.78,0.0){Target $N$}
  \end{picture}
  \end{center}
\caption{The alignment result for a long dress.}
\label{fig:dress}
\end{figure}
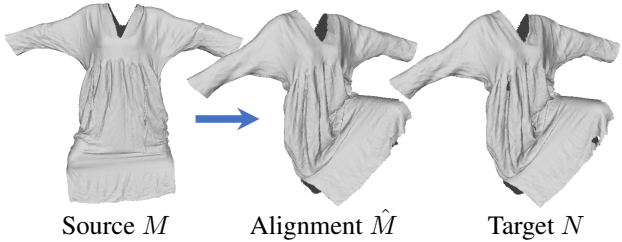

%% file: figs/tbl2-singlecol.tex
\begin{table}
\begin{center}
\begin{tabular}{|l|c|c|c|}
\hline
\multirow{2}*{Method} & Chamfer Dist. & LPIPS-VGG & SSIM \\
   & ($\times10^{-3}$) $\downarrow$ & ($\times10^{-2}$) $\downarrow$ & $\uparrow$\\
\hline
Ours               & \textbf{2.780} & \textbf{8.866} & \textbf{0.824} \\
DVO-S1          & 2.899 &10.868 & 0.813 \\
CoordNet-S1 & 2.793  & 9.188 & \underline{0.821} \\
LR-S2              & 2.834 & 9.146 & \underline{0.821} \\
NR-S2              & 2.897 & 9.344 & 0.819 \\
NoTex              & \underline{2.783} & \underline{9.097} & \underline{0.821} \\
\hline
\end{tabular}
\end{center}
\caption{Quantitative results of ablation studies. We report the metrics for S02 (knee-long coat). The best results are in boldface and the second best results are underlined.}
\label{tbl:abl-separate}
\end{table}

%% file: sections/sec5_conclusion.tex
\section{Conclusion and Future Work}

In this work, we introduce a two-stage pipeline that aligns garments by utilizing intrinsic manifold properties and neural deformation fields. Our intrinsic deformation network for 3D fitting leverages manifold continuity instead of extrinsic 3D continuity, and can thus handle difficult garment types such as long coats, as well as being robust to poor initializations of the base template. Furthermore, our method extends the Functional Maps framework~\cite{ovsjanikov2012fm} by introducing non-linearity with neural deformation fields, achieving texture-level high accuracy garment alignment where highly non-rigid and non-isometric deformations are present.

\paragraph{Limitations and Failure Cases}

As discussed in \cite{eisenberger2020smoothshells}, extrinsic methods are often more suitable for shapes with inconsistent topology. While we have also relied on extrinsic (3D) fitting in the coarse stage, our final alignment comes from intrinsic embeddings, which may not be suitable if the target shape is topologically different from the source shape. Moreover, since it is not intuitive how physics-based constraints can be added to the intrinsic space, the output may not always be physically correct. Another limitation is that in both stages we model the alignment between complete shapes that are nicely segmented without noise and self-occlusion. Noisy scans or partial scans, which are very common for garment captures, cannot be perfectly handled yet. Currently, our method does not explicitly handle noise but can still produce a reasonable alignment for noisy data (Fig.~\ref{fig:noise}). If the target shape has deep wrinkles (which is essentially a type of partiality due to self-occlusion), the vertices near the wrinkle would appear smoothly squeezed together as shown in Fig.~\ref{fig:deepwrinkle}.

\paragraph{Future Work}

Aligned garment data enable a number of applications, \eg, texture capture or physical parameter estimation. Our method can also be used as a tool for building correspondences for garment datasets to be used for learning tasks, \eg, learning deformations and textures of garments to drape them on human avatars, or learning shape descriptors specific to garments for correspondence inference. Efforts should also be devoted to improve various aspect of our method. In our current setting it is assumed that garment data are nicely segmented scans, without noise and occlusion during capture. Future work should extend this pipeline to support imperfections such as noise, partiality and segmentation failures. Generalizing the LBO eigenfunction computation to point clouds can also make the pipeline more compatible with depth sensor-based captures.

\input{figs/fig7_tex}
\input{figs/fig8-rebuttal-deepwrinkle_tex}

%% file: figs/fig7_tex.tex
\setlength{\unitlength}{\linewidth}
\begin{figure}
  \begin{center}
  \begin{picture}(1,0.39)(0,0)
  \put(0,0.03){\includegraphics[width=\linewidth]{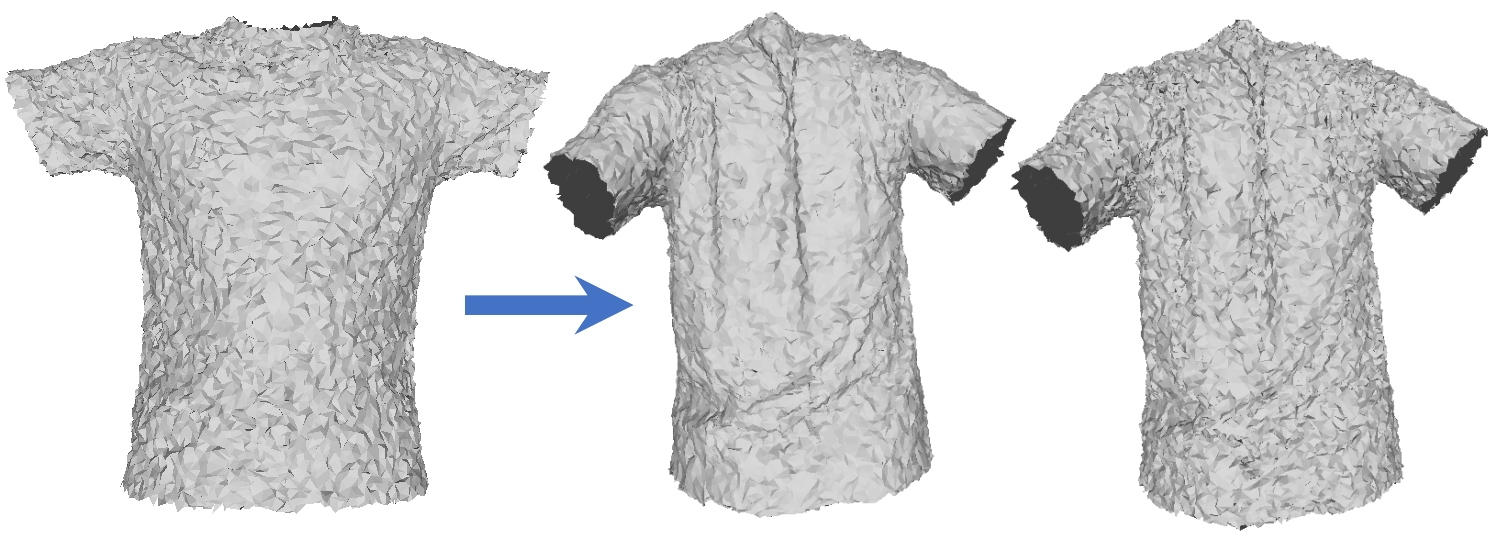}}
  \put(0.04,0.0){Noisy source $M$}
  \put(0.4,0.0){Alignment $\hat M$}
  \put(0.72,0.0){Noisy target $N$}
  \end{picture}
  \end{center}
\caption{The alignment result for data with synthetic noise.}
\label{fig:noise}
\end{figure}
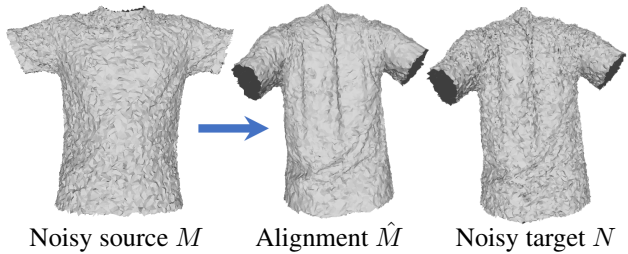

%% file: figs/fig8-rebuttal-deepwrinkle_tex.tex
\setlength{\unitlength}{\linewidth}
\begin{figure}
  \begin{center}
  \begin{picture}(1,0.44)(0,0)
  \put(0,0.04){\includegraphics[width=\linewidth]{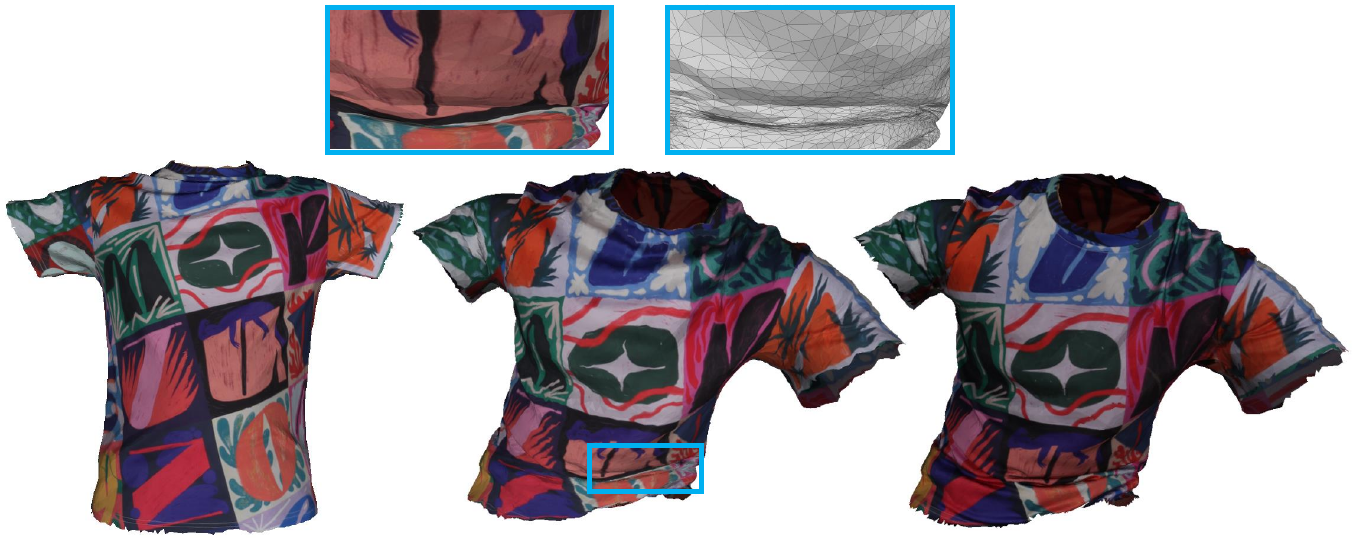}}
  \put(0.06,0.0){Source $M$}
  \put(0.34,0.0){Alignment $\hat M$}
  \put(0.7,0.0){Target $N$}
  \end{picture}
  \end{center}
\caption{The alignment result for data with deep wrinkles.}
\label{fig:deepwrinkle}
\end{figure}
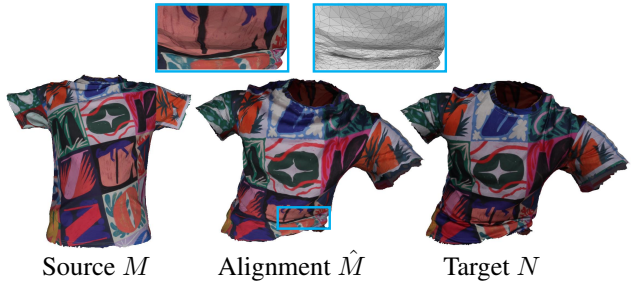